%% file: main.tex
\documentclass[10pt,a4paper,twocolumn]{article}
\usepackage{bachelorproject}
\usepackage[activate={true,nocompatibility},final,tracking=true,kerning=true,spacing=true,factor=1100,stretch=10,shrink=10]{microtype}
\usepackage{amsthm}
\usepackage{capt-of}
\usepackage{makecell}

\makeatletter
\newcommand{\thickhline}{%
    \noalign {\ifnum 0=`}\fi \hrule height 1pt
    \futurelet \reserved@a \@xhline
}
\newcolumntype{"}{@{\hskip\tabcolsep\vrule width 1pt\hskip\tabcolsep}}
\makeatother
\title{\vspace{-0.5in}{\bfseries 
    Investigation on the generalization of the\\ Sampled Policy Gradient algorithm}\\
}

\author{
  Stolt Ans\'o, Nil\\
  \texttt{nilstoltanso@gmail.com}
}

\date{\vspace{-5ex}}

\begin{document}

\twocolumn[
  \maketitle
  \begin{@twocolumnfalse}
    \input{./tex/abstract.tex}
  \end{@twocolumnfalse}
]

\thispagestyle{firststyle}
\setcitestyle{square}
\bibliographystyle{unsrt}

\input{tex/introduction.tex}
\input{tex/game.tex}
\input{tex/rl.tex}
\input{tex/setup.tex}
\input{tex/results.tex}
\input{tex/conclusion.tex}
\input{tex/acknowledgements.tex}

\bibliographystyle{plain}
\bibliography{main}

\clearpage
    \begin{appendices}
        \input{tex/appendix.tex}
    \end{appendices}

\end{document}

%% file: tex/abstract.tex
\begin{abstract}
The Sampled Policy Gradient (SPG) algorithm is a new offline actor-critic variant that samples in the action space to approximate the policy gradient. It does so by using the critic to evaluate the sampled actions. SPG offers theoretical promise over similar algorithms such as DPG as it searches the action-Q-value space independently of the local gradient, enabling it to avoid local minima. This paper aims to compare SPG to two similar actor-critic algorithms, CACLA and DPG. The comparison is made across two different environments, two different network architectures, as well as training on on-policy transitions in contrast to using an experience buffer. Results seem to show that although SPG does often not perform the worst, it doesn't always match the performance of the best performing algorithm at a particular task. Further experiments are required to get a better estimate of the qualities of SPG.
{ \vspace{6ex}}

\end{abstract}

%% file: tex/introduction.tex
\section{Introduction}\label{sec:introduction}
Reinforcement learning is a machine learning paradigm in which agents learn what actions to take in an environment in order to maximize a reward function. The reward function is often inherent to the environment and acts as a supervision signal~\cite{RL:AI}. Training commonly takes place in simulated environments, as the levels of noise can be controlled and all information in the system can be accessed. Being able to speed up a simulation also leads to a more time-efficient training process.

Traditionally, algorithms such as Q-learning \cite{Q-Learning} have shown to be well suited at control tasks. With the recent popularity of artificial neural network (ANN), a paradigm shift has lead the field to move away from hand-designed features. The use of ANNs as function approximators has helped solve control tasks with ever-higher dimensional observation spaces. Tesauro was among the first to show the effectivity of such an approach by showing that a multi-layer perceptron (MLP) could learn to play Backgammon~\cite{tesauro1995}. More recently, the generality of deep Q-networks (DQN) was popularized by DeepMind after solving a wide range of tasks at the pixel level~\cite{mnih2015}. However, many of these approaches can only handle discrete action spaces.

A well known family of reinforcement learning algorithms are the actor-critic methods. An actor-critic architecture is made up of an actor, which acts out a parameterized policy $\pi_\theta$, and a critic that aims to learn the expected discounted return provided by the environment \cite{Sutton1999}. The benefit of modelling the policy explicitly this way is that you can map states into a continuous action-space directly. There exist many ways to train an actor, but regardless of the approach, the actor relies on a gradient signal provided by the critic to improve its policy. Critics can be broken down into two categories: ones that predict the expected return of a given state $s_t$ under the current policy $\pi$, and ones that predict the return of taking action $a_t$ in a given state $s_t$ with thereafter following of policy $\pi$. The prior is termed the value function $v(s_t)$, while the latter is termed the action-value function $q(s_t, a_t)$. 

A well known actor-critic algorithm is Deterministic Policy Gradient (DPG) \cite{DPG}. DPG is an off-policy algorithm that has an action-value function for a critic. The actor is trained through the use of the policy gradient $\nabla_{\theta} E \left[J(s)\right]$ as given by the critic. Another approach to training the actor is that of the Continuous Actor Critic Learning Automaton (CACLA) algorithm \cite{CACLA}. CACLA uses the sign of the temporal difference (TD) error to determine whether the actor should be updated during a training step. Only if the TD-error produced by the noisy action is positive, the policy is updated to make that action more likely. Finally, the Sampled Policy Gradient (SPG) \cite{SPG} algorithm takes the actor's update procedure one step further by sampling from the action-space an arbitrary amount of times. The critic is an action-value function. By allowing the critic to determine the action out of the sample predicted to yield the highest return, the actor can use it to move the policy in that direction.

\subsection{Previous research}

The field of reinforcement learning has seen great broadening of its applications in recent years thanks to some of the advances in deep learning. The use of tabular methods to map states into expected rewards has grown out of use due to its limitations in memory requirements as the state-space complexity of our problems rise. Furthermore, although introducing handcrafted features can help reduce the problem's complexity, these might not generalize to other environments and may even introduce human bias. Given a large enough hidden layer, an ANN has been shown to be able to approximate any continuous function on the input space to any degree of accuracy \cite{Cybenko1989}. ANNs thus offer a method of mapping very large state-spaces to the expected rewards of each action. With the use of convolutional neural networks (CNNs), this approach has been shown to achieve good performance by relying solely on pixel input. One of the first to show the generality and robustness of this approach was Deepmind with their DQN architecture \cite{mnih2013}.

Discretizing the action-space brings a set of limitations however. As the number of degrees of freedom in the action-space increases, the number of possible actions increases exponentially. Sometimes, one can get away with discretization of a continuous action-space \cite{DeepAgar}, but such approaches often throw away information about the inherent structure of the action domain that may be essential to the task. This is for example the case in some areas of robotics where one would like fine-grained control over its actuators. Actor-critic architectures allow the mapping of the state-space directly to the action-space. More specifically, the actor is responsible for providing real-valued actions when presented with a state. 
Perhaps one of the simpler actor-critic algorithm is CACLA \cite{CACLA}. CACLA has a value function $v(s_t)$ as a critic that predicts the expected discounted return of a given state $s_t$ under the current policy $\pi$. This makes CACLA an on-policy algorithm. The actor is trained by analysing whether a state-action pair yielded a higher-than-predicted value. If that is the case, the action will be reinforced so as to be more likely predicted by the actor in the future.

A much more well-known algorithm is DPG \cite{DPG}. DPG has a Q-value function $q(s_t,a_t)$ for a critic which, given an action alongside a state, predicts the expected return of taking action $a_t$ in state $s_t$, with thereafter following of policy $\pi$. The actor is trained using the critic's estimation of the action-space's gradient direction. The gradient is used to change the actor's parameters in a way that will maximize the Q-value predicted by the critic. DPG is thus not dependent on transitions for training the actor as it only requires a valid state of the environment, making it an off-policy algorithm.

Lastly, Sampled Policy Gradient (SPG) is an algorithm that samples the action-space in order to approximate the direction of the global maximum~\cite{SPG}. SPG relates to the policy gradient theorem in the sense that it approximates the term $\nabla_{\theta} Q(s,\pi(s))$ through sampling. The algorithm has a Q-value function $q(s_t,a_t)$ for a critic, but similar to CACLA, the actor is updated by applying noise to an action. Given a state, the actor predicts an action. Through the use of noise, the action-space around the predicted action is sampled. These samples are presented to the critic, and the action with the best predicted Q-value is used to update the actor. Unlike CACLA, since the critic directly learns the returns of action-state combinations, the likelihood of actions under the policy that generated the transitions has no effect on the return predictions, making SPG an off-policy algorithm. This also means that given a good critic, a policy can be trained off-line without the need to interact with the environment. Furthermore, the theoretical appeal of this algorithm comes from not directly relying on the local gradient (as DPG does), which can help escape local minima.

In recent years, other algorithms have been developed that explore the idea of improving a policy through some kind of local search. The stochastic policy gradient family of algorithms have a policy that predicts a distribution of actions, where the action is chosen by sampling from said distribution. A notable algorithm belonging to such family is A3C \cite{A3C}. A problem that arises with such methods is that it is hard to ensure that an action sampled far from your distribution center is truly better, as it could be the case that the prediction is better accidentally due to lack of previous training transitions in that region of the state-action space. Some algorithms mitigate this issue by attempting to constrain an update to a `safe region' around the current policy, so as to discourage sudden shifts of the policy that are too large. Trust Region Policy Optimization (TRPO)~\cite{TRPO} and Proximal Policy Optimization (PPO)~\cite{PPO} are among the most famous algorithms of this kind. Despite what these approaches gain in performance, the algorithms tend to require to be trained on-policy and their update functions can end up being quite complex.

\subsection{Contributions of this paper}
Given that SPG has only so far been examined in one single specific domain in the past, this research aims to investigate how its performance generalises to different types of tasks one might apply such algorithm to. The research will focus on comparing SPG to DPG and CACLA. This will be done so in two different environments. The first will the original game SPG was tested on, Agar.io, in order to ensure reproduceability of the results. The second is the HalfCheetah-v2 robotics tasks in the MuJoCo physics simulation. 

More importantly, this research investigates how SPG performs in different versions of the aforementioned tasks. One task is going to be learning from on-policy experiences in both of the environments. In other words, no experience replay will be used and instead, $n$ separate workers will each use the current policy to gather experiences from $n$ parallel synchronized simulations. The experiences will be discarded after having trained on them once. The other task will involve using a CNN instead of a regular MLP in order to shed some light on how SPG might generalise to learning from environments where only pixels are available and deep networks are required.

\subsection{Paper outline}
This paper has the following structure. Section 2 will describe the core behaviours that each environment requires. This will go over the reward functions used, what an episode in the environment is like, what the state representation for each environement was, and the intricacies of the multi-worker parallelization of the environment. Section 3 will cover core concepts of reinforcement learning that were used in this research, including a detailed description of the SPG algorithm. Next, section 4 will contain the experimental setup used. This will be followed by an outline of the results obtained in section 5. Finally, section 6 will provide discussion and evaluation of the results, as well as an outline for future research.

%% file: tex/game.tex
\section{Environments and Tasks}\label{sec:game}
As mentioned in the previous section, the algorithms were trained on two different environments: Agar.io, a game involving making the player's cell eat pellets to grow, and HalfCheetah-v2, a task involving making a two dimensional cheetah learn to walk. Each of these environments had two tasks: training with the use of an experience replay buffer and training on on-policy experiences without the use of a buffer. Additionally, we also train on a task in Agar.io involving learning from pixels which we aim to compare to the results of the regular Agar.io task.

The implementation of the experience buffer is rather straight forward. One single agent is acting on one single environment. Every transition is stored in the buffer in the form of a tuple containing the state $s_t$, the action taken $a_t$, the reward received $r_t$, the new state $s_{t+1}$, and a boolean indicating whether the episode ended in this transition. When the buffer is full, a new transition replaces the oldest transition in the buffer. A training step involves sampling a batch of experiences from the buffer at random with uniform probability.

The task of training with on-policy experiences does not involve an experience buffer. Parallel simulations of the environment are run in a synchronized manner. First, so as to gather experiences that are distributed across different time steps of the episode, we have each simulation advance by a different number of time steps. The number of time steps is given by:
\begin{equation*}
   \text{worker id} \cdot\frac{\text{episode length}}{\text{number of workers}}
\end{equation*}
where \textit{worker id} is a unique number assigned to every simulation ranging from 0 up to the total number of parallel simulations. The transitions being generated during this process are discarded. Once all simulations are advanced, the simulations proceed one step at a time in a synchronized manner collectively providing the training process with a batch of transitions.

\subsection{Agar.io}
Agar.io is an online game in which the player controls one or more cells. The game has a top-down perspective where the size of the visible area to the player is based on the mass and count of their cells. The goal of the game is to grow as much as possible. This can be done by having the player's cell absorb food pellets, viruses, or smaller enemy player's cells. For the purpose of this research, a simplified version of the task is used, where the player only ever has a single cell centered on the middle of the screen and the only food source are scattered pellets. We name this task the `pellet collection task'. We make the change in mass the reward function of this environment. Every cell in the game loses a small percentage of its mass in every time step. This makes it harder for large cells to grow quickly and it punishes inaction or hesitation.

\begin{figure}[h]
    \centering
    \includegraphics[width=.45\textwidth]{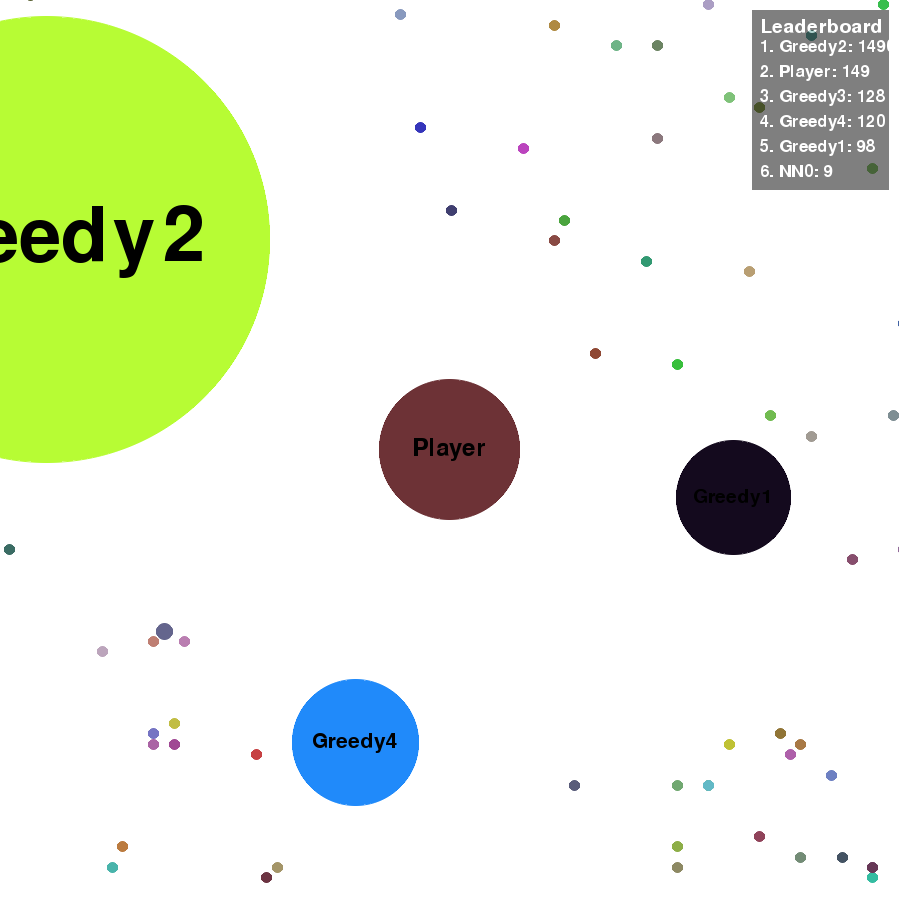}
    \caption{A clone of the game Agar.io used for this research. The player has one cell in the center of the screen. The little colored dots scattered across the screen are pellets that can be consumed to grow in mass. Although not present in the pellet collection task, the other cells present are enemy players. These are shown in this figure to give a perspective of the range of sizes that the player can take.}
    \label{fig:agario}
\end{figure}

The game has simple controls. The cursor's position on the screen determines the direction all of the player's cells move towards. A cursor location of (0,0) thus means the cells will move towards the top-left corner, while a cursor position of (1,1) means the cells will move towards the bottom-right corner. Figure \ref{fig:agario} shows a screenshot of the clone of Agar.io used for this research. Because this task has a high frame-rate relative to meaningful events occurring, we introduce frame-skipping \cite{mnih2013}. In frame-skipping, a set number of frames are skipped after having chosen an action. The action is applied across all the skipped frames and the reward of these transitions are added up. Rewards are therefore larger and quickly propagated in state-action space, which leads to successive states being more different from each other. A frame-skip rate of 7 frames is used for this environment, which leads to one action decision being taken every 8th frame. The episode length is 20,000 frames, which given frame-skipping translates to a total of 2,500 state-action transitions. Frame-skip rate is only used for the Agar.io tasks, and as such, the tasks regarding the HalfCheetah-v2 environment will not make use of it.

The main state representation used for this task is that of vision grids. Vision grids help overcome the issue of large state spaces by extracting features and grouping them into areas in a grid-like fashion. Despite the features having to be hand-crafted, the approach helps filter irrelevant information for the agent. Such approaches have been employed in games such as Starcraft \cite{shantia2011}, Tron \cite{knegt2017}, and more recently Dota2 \cite{OpenAIFive}.

\begin{figure}[!ht]
    \centering
    \includegraphics[width=.45\textwidth]{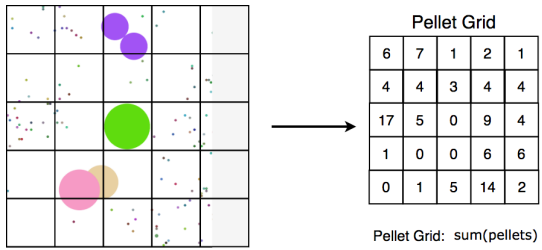}
    \caption{Visualization of the the pellet vision grid. Each cell sums up the mass of pellets present in it.}
    \label{fig:vision_grid}
\end{figure}

For this task, we use a single 11 by 11 vision grid, where each individual area unit provides a floating-point number corresponding to the sum of pellet mass present in it (see Figure \ref{fig:vision_grid}). The array of values is then flattened, after which the current mass and view size of the player are concatenated to it. The resulting state representation is a vector of 123 floating-point numbers.

One additional variation of the pellet collection task investigated in this research is one where the agent has to learn from the 42 by 42 gray-scale pixel representation of the task. For this, a CNN network will be used as discussed in section 4.

The exact algorithm and environment implementations used for this research can be found at \href{https://github.com/NILOIDE/A.I.gar}{\small{https://github.com/NILOIDE/A.I.gar}}. The Agar.io environment used has since been wrapped into an OpenAI gym environment (thanks to Anton Wiehe) and can be found at \href{https://github.com/NotNANtoN/gym\_aigar}{\small{https://github.com/NotNANtoN/gym\_aigar}}.

\subsection{HalfCheetah-v2}
HalfCheetah-v2 is a robotics task provided by OpenAI's gym toolkit \cite{OpenAIgym}. The task runs on the MuJoCo physics simulator. In the task, the agent is a two-dimensional cheetah controlled by seven actuators. The state representation corresponds to the position and velocity of the cheetah, current angle and angular velocity of its joints as well as the torque on the actuators. The reward function in this environment is a mixture between the amount of contact the limbs have on the floor and the displacement of the agent from the starting position.

\begin{figure}[!ht]
    \centering
    \includegraphics[width=.45\textwidth]{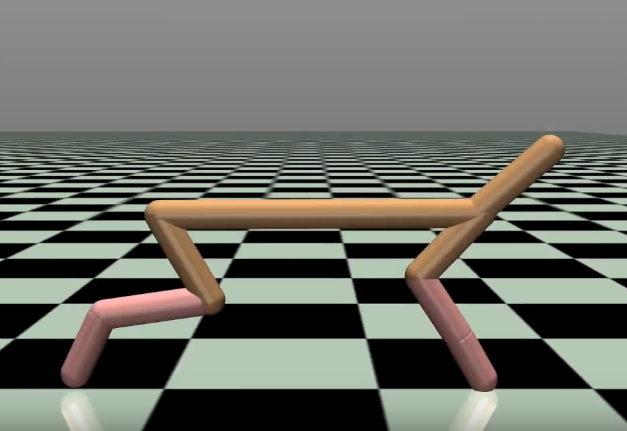}
    \caption{The HalfCheetah-v2 environment use for this research. The agent is a two-dimensional cheetah with the goal of moving to the right.}
    \label{fig:halfcheetah}
\end{figure}

%% file: tex/rl.tex
\section{Reinforcement Learning}\label{sec:rl}

This paper follows the standard reinforcement learning (RL) conventions \cite{RL:AI} of modelling the agent's interaction with the environment as discrete discrete timesteps. Each timestep $t$, the agent is presented with an observation $x_t$, which it uses to base its next action $a_t$, after which a scalar reward $r_t$ is provided. The environment can thus be viewed as a Markov Decision Process (MDP), where the agent can transition from a state to a new state by taking an action. The Markov property is assumed across our environments; stochastic transition probabilities between states are only dependent on the current state and action.

\subsection{Exploration}
Two forms of exploration are used in this research. The first is a simple optimistic initialization of the critic's predicted return. This leads to natural initial exploration \cite{RL:AI}. The second and most important form of exploration is the addition of Gaussian noise to the action being taken by the actor. This forces the agent to explore alternative regions of the action-space over the course of training. The Gaussian noise is always centered on the predicted action and starts with an initial standard deviation of 1.0. The noise is then exponentially decayed down with a half-life of 0.05 according to the following formula:
\begin{equation*}
    N(t)=N_{0} e^{-\lambda t}
\end{equation*}
Where $t$ is the given time step, $N_{0}$ is the standard deviation at $t=9$, and $\lambda$ is the decay coefficient which is determined by setting $t=\frac{1}{2}t_{max}$ and $N(t)$ to the half-life value (0.05 in our case).

\subsection{Target Networks}
The combination of function approximators, bootstrapping of return estimates, and off-policy training can lead to a positive feedback loop termed the deadly triad \cite{RL:AI}. This has the danger of leading to instability and divergence. To stabilize deep Q-networks, Mnih et al. introduced target networks~\cite{mnih2013}. This approach involves copying the parameters of our networks into a new network which remains static. The target network is used for prediction, and only updated after a certain number of steps. This ensures prediction is done by a slightly different network, which mitigates the deadly triad's effect. Target networks were used both for the actor and the critic.

\subsection{Sampled Policy Gradient (SPG)}
The SPG algorithm uses the same network architecture as DPG would. The main difference comes in the update procedure of the policy's parameters. DPG trains the actor through the application of the chain rule to the expected return from the start distribution $J(\theta)$ with respect to the actor parameters $\theta$ \cite{DDPG}. Instead, SPG takes an approach similar to CACLA's by sampling the action-space around the actor's prediction. A batch of actions are sampled, and if the best sampled action is predicted by the critic to have a higher Q-value than the policy's current estimate, that action is made more likely. The target during the actor's update step can be summarized as follows:
\begin{equation}
    Q\left(s_{\mathrm{t}}, a_{\mathrm{t}}\right)>Q\left(s_{t}, \pi\left(s_{\mathrm{t}}\right)\right) : \operatorname{Target}\left(s_{t}\right)=a_{\mathrm{t}}
\end{equation}
where $a_t$ is the sampled action with highest Q-value selected by the critic.

%% file: tex/setup.tex
\section{Experimental Setup}\label{sec:contents}
\subsection{General setup}
Each algorithm received the same amount of training steps in each experiment. The number of training steps and the episode length used for the testing varies across the environment used. At each 5\% of the training, a copy of the network is saved to obtain the testing performance of the algorithm. This results on 21 copies saved, including the network with random weights prior to training. The training procedure was performed 10 times so as to obtain an average measure. Furthermore, the testing procedure at each 5\% for an individual training run was replicated 5 times. Given that a testing round was one episode long, a data point at each 5\% of training is thus the average of 50 separate episodes.

The total number of training and testing steps were the following:
\begin{enumerate}
    \item \textbf{Agar.io vision grid:} The total amount of training steps used in this task were 500,000 state transitions. An episode's length in this environment was 20,000 frames. Due to skipping 7 frames for every action taken, one state transition accounts for 8 frames. This means that an episode amounts to 2500 state transitions. During testing, the agent was allowed to perform for one episode without noise being applied to its actions. The start of an episode places the agent at a random position in the map and randomizes pellet spawning positions. The seed for each training run is also randomized.
    
    \item \textbf{Agar.io pixel input:} The episode length and testing length were the same as in the Agar.io task above, but the training steps were 300,000.
    
    \item \textbf{HalfCheetah-v2:} The total amount of training steps were 200,000 state transitions. An episode's length in this environment was 1,000 state transitions. During testing, the agent was allowed to perform for one episode without noise being applied to its actions.
\end{enumerate}

To incentivize exploration, Gaussian noise was applied to the action performed by the policy. This was done so regardless of the algorithm. The standard deviation (SD) of the Gaussian noise decayed over time in an exponential manner, plateauing to zero towards the end of the training procedure. The starting SD was 1.0 and was made to have decreased to 0.05 after 50\% of training steps.

\subsection{Task variations}
The Agar.io vision grid task and the HalfCheetah-v2 task were ran under two different variations. One variation involved the use of an experience buffer, which allows training on previous transitions. The other variation was without the use of an experience buffer, which forces the algorithms to learn from transitions obtained under the current policy. The Agar.io task involving the use of pixels as input was only tested on the experience replay task, as the on-policy task's computational requirements were too high.

\begin{enumerate}
    \item \textbf{Experience replay task:} As the name entails, in this task an experience replay buffer was used. A single agent gathered experiences in one environment and each experience was placed into the buffer. Once full, the experiences in the buffer were replaced in a first-in-first-out basis. At each training step, 32 experiences were sampled with a uniform probability from the buffer in order to perform mini-batch gradient descent on the networks. The experience buffer sizes for each environment were empirically set to the following:
    \begin{itemize}
        \item \textbf{Agar.io vision grid: } 40,000.
        \item \textbf{Agar.io pixel input: } 20,000.
        \item \textbf{HalfCheetah-v2: } 15,000.
    \end{itemize}
    
    \item \textbf{On-policy experiences:}
    For this task, 32 workers using the same policy parameters were individually placed in 32 separate environments running in separate processes. Each environment was initialized under a separate seed. No experience buffer was used in this task. Instead, after each step, the transitions performed by each agent were collected into a batch that was sent to a separate process in charge of performing a training step on the global copy of the network. After each training step, the workers updated their parameters with those of the global network. All of this happened in a synchronous manner.
    
\end{enumerate}

\subsection{Network structures}
Neural networks were used as function approximators for the actor and the critic. All parameters in both networks were randomly initialized under a Glorot uniform distribution. After each state transition performed, both networks were trained with a mini-batch. The prediction during state transitions was performed by target networks, which are separate copies of the actor and the critic. These copies were updated every 1,500 steps. The network structure used for each environment were the following:

\begin{enumerate}
    \item \textbf{Agar.io vision grid:} The network consisted of two fully-connected hidden layers of 100 rectified linear units each. The critic had one output node, while the actor had 2 output nodes (one for the x-direction and one for y-direction). The critic used a linear output, while the actor had sigmoid non-linearities applied to the ouput in order to constrain the action in the range (0, 1).  The state representation being used was a 11 by 11 grid. The flattened grid, together with the mass and the view size resulted in an vector input of size 123. The input size to the actor was thus 123, while the critic input was of size 125 (due to the action being of size 2). The total amount of parameters were 22600 for the actor. The Q-value critic in SPG and DPG had 22500 parameters, while the critic in CACLA had 22,601.
    
    \item \textbf{Agar.io pixel input:} The network consisted of two convolutional layers, followed by a fully connected layer. The parameters of the convolutional layers were shared by the actor and the critic. The first convolutional layer used 32 filters of kernel size 8 and stride 4, followed by a ReLU activation function. The second convolutional layer used 64 filters of kernel size 4 and stride 2, followed by a ReLU activation function. The hidden state was subsequently flattened and fed to a fully connected layer of 100 rectified units. One more fully connected layer of 100 rectified units followed. Unlike in the `vision grid` environment, the mass and view size values were not appended. The actor had 2 output units that used a sigmoid to constrain the output in the range (0, 1), while the critic had 1 linear unit. The input image was a 42 by 42 grayscale image to both networks. In addition to the input image, the critic appended to the actor's action to the flattened output of the convolutional layers prior to the forward pass of the fully-connected layer. The actor's total parameters were 102,914. The Q-value critic in SPG and DPG had 103,013 parameters, while CACLA's critic had 102,813. Note that some of the parameters in the actor and the critic are shared due to sharing convolutional layers.
    
    \item \textbf{HalfCheetah-v2:} The architecture used for this environment was made up of two fully connected layers of 100 rectified linear units each. The actor this time had 6 output nodes (one for each joint) and the state space was of length 17. The total number of parameters was 12,506 for the actor. The Q-value critic had 12,601 parameters in SPG and DPG, while the critic in CACLA had 12,001.
    
\end{enumerate}

The optimizer used to train the networks was Adam \cite{adam}. Hyperparameters involving learning rates were determined through a grid-search approach. The starting hyperparameters during the grid-search for individual algorithms were those found in their original papers \cite{DPG} \cite{CACLA} \cite{SPG}, and then optimized on the Agar.io vision grid task. The exact hyperparameters used can be found in the appendix. 

%% file: tex/results.tex
\section{Results}\label{sec:layout}
\subsection{Experimental results}
\begin{figure}[!ht]
    \centering
    \includegraphics[width=.45\textwidth]{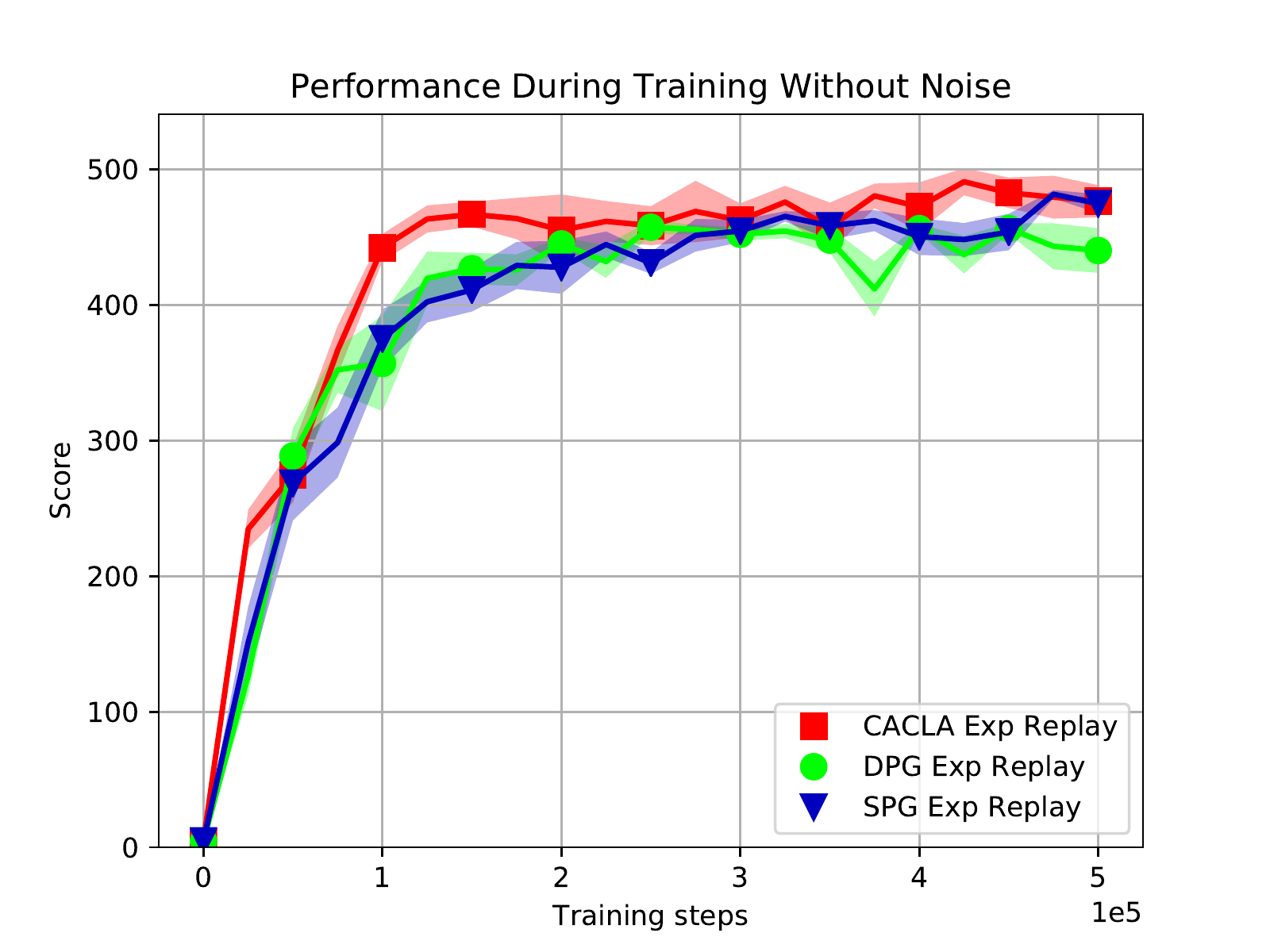}
    \caption{During-training performance in the task involving Agar.io with vision grids when using an experience buffer. Each point represents the average of the 10 testing rounds and the shaded area denotes its 1 S.D. range. Results are averaged over 10 simulations.}
    \label{fig:ER_Agar_results}
\end{figure}

The during-training performances of the three algorithms in the Agar.io vision grid environment under the experience buffer task appear to replicate the results obtained in \cite{SPG}. These can be seen in Figure \ref{fig:ER_Agar_results}. All three algorithms have around the same performance at convergence (450-500 mass) and all converge at approximately the same speed.

\begin{figure}[!ht]
    \centering
    \includegraphics[width=.45\textwidth]{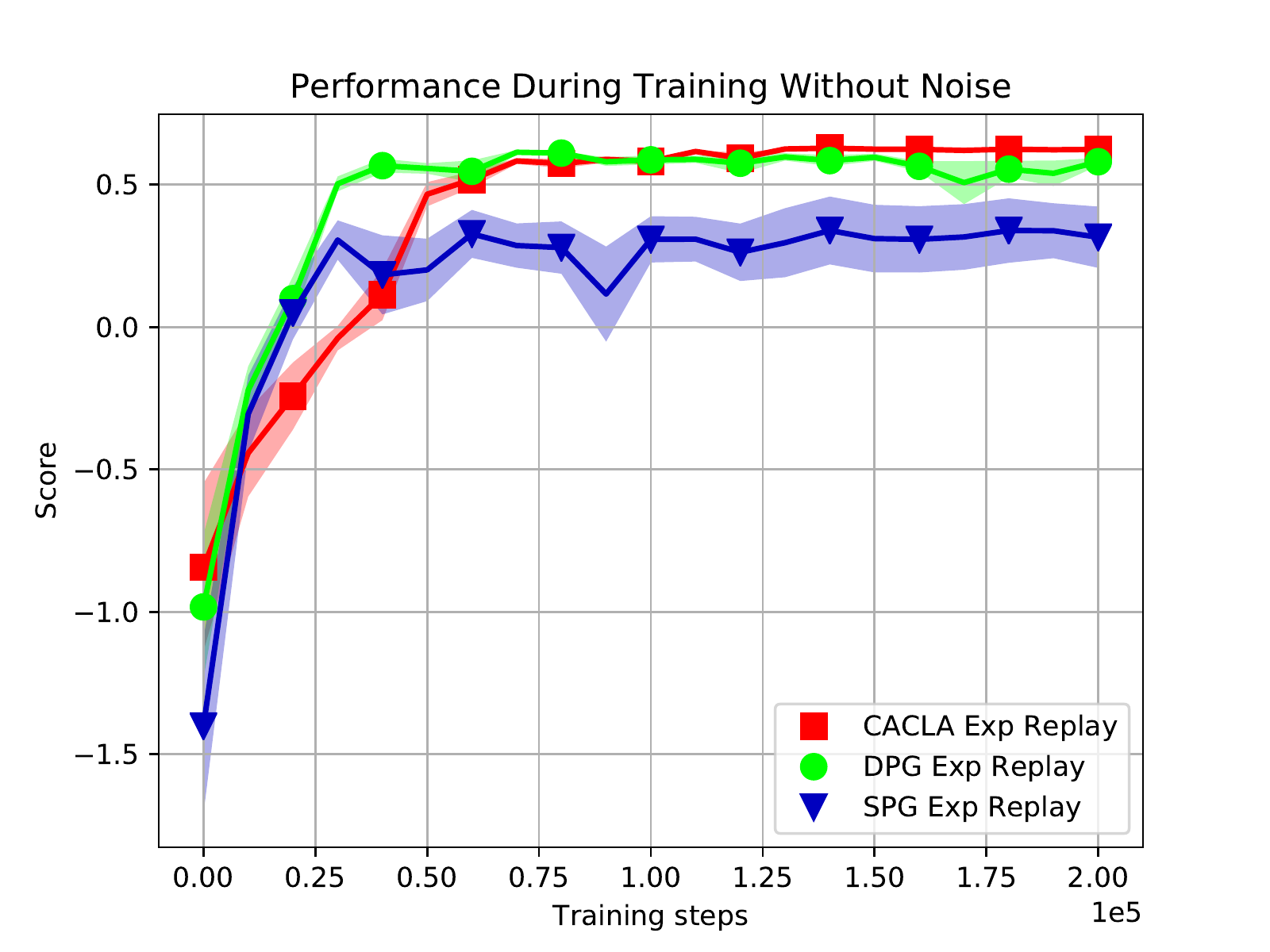}
    \caption{During-training performance in the task involving HalfCheetah-v2 when using an experience buffer. Each point represents the average of the 10 testing rounds and the shaded area denotes its 1 S.D. range. Results are averaged over 10 simulations.}
    \label{fig:ER_HC_results}
\end{figure}

The same cannot be said about the HalfCheetah-v2 environment when using an experience buffer (Figure \ref{fig:ER_HC_results}). While CACLA and DPG have a similar performance of 0.6, SPG appears to converge at a slightly lower average score of 0.3. This might be explained by the higher standard deviation of SPG, which might indicate that although some of the runs reached comparable performances to the other algorithms, other runs had a low score. 

\begin{figure}[!ht]
    \centering
    \includegraphics[width=.45\textwidth]{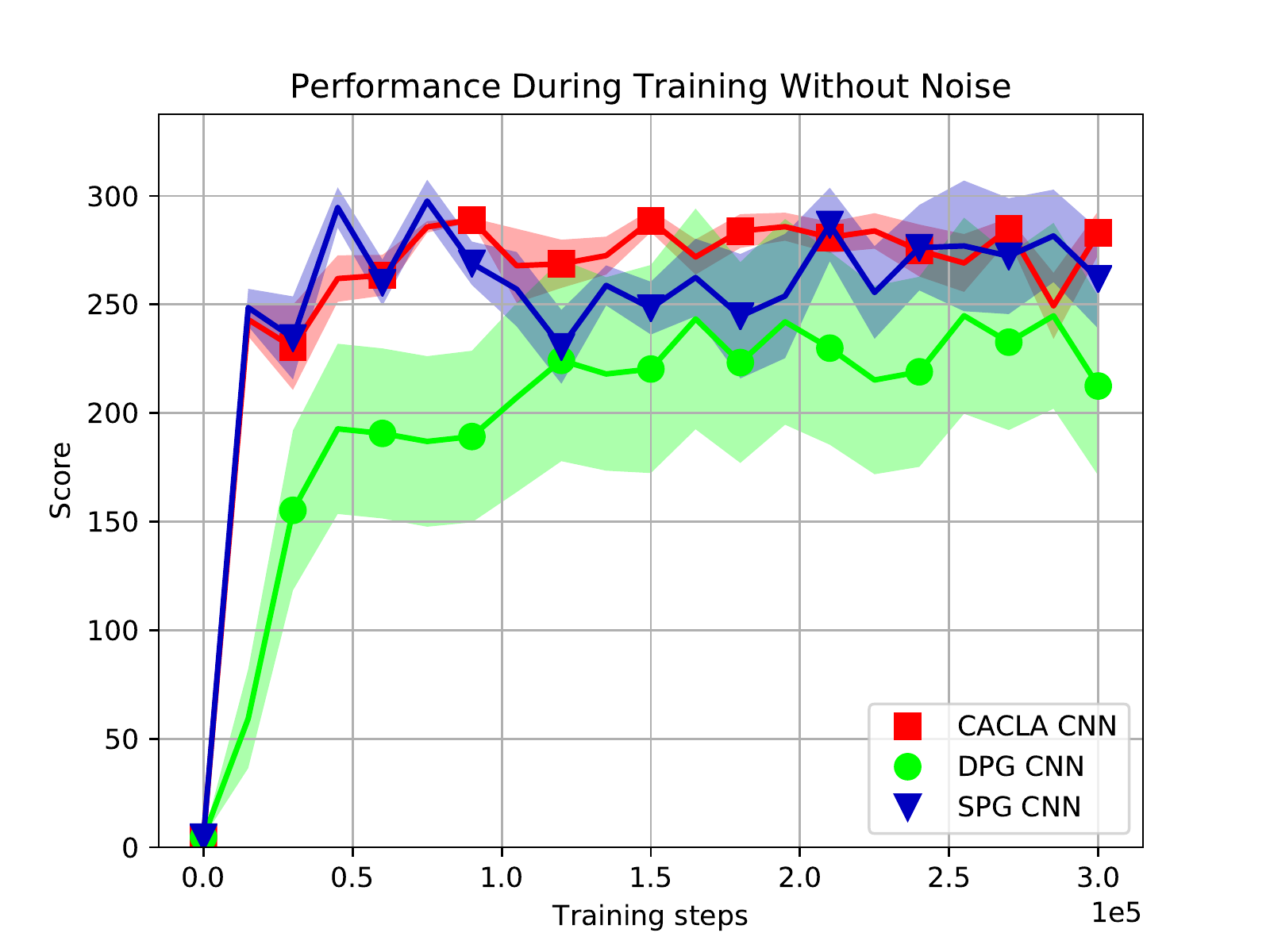}
    \caption{During-training performance in the task involving Agar.io when learning from pixels. Each point represents the average of the 10 testing rounds and the shaded area denotes its 1 S.D. range. Results are averaged over 10 simulations.}
    \label{fig:CNN_Agar_results}
\end{figure}

In the Agar.io pixel input task (Figure \ref{fig:CNN_Agar_results}), the performances of the algorithms appear to differ from those seen in the environment with vision grid input (Figure \ref{fig:ER_Agar_results}), despite the algorithms having access to an experience buffer in both cases. DPG appears to have a lower average score than the other two algorithms, although its standard deviation is much higher. In two of the 10 training runs of DPG, the algorithm appeared to have completely failed to learn and appeared to have developed performances comparable networks with random weights. In the two failed runs, after an initial increase in performance, DPG converged down to a mass 0 after 50,000 training steps, bringing the overall average down significantly.

\begin{figure}[!ht]
    \centering
    \includegraphics[width=.45\textwidth]{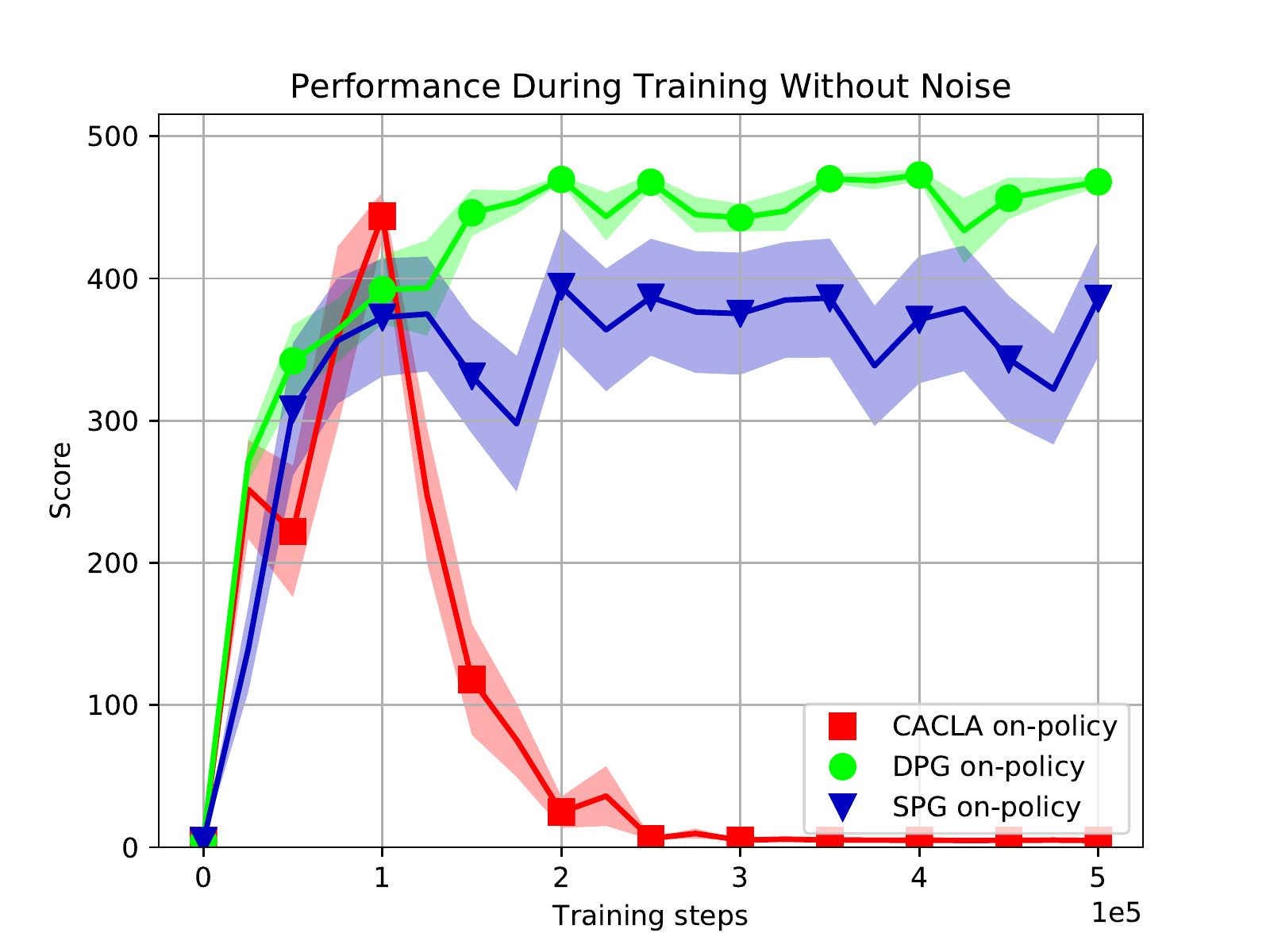}
    \caption{During-training performance in the task involving Agar.io with vision grids without an experience buffer. Each point represents the average of the 10 testing rounds and the shaded area denotes its 1 S.D. range. Results are averaged over 10 simulations.}
    \label{fig:OP_Agar_results}
\end{figure}

The Agar.io vision grid environment under the on-policy training task (Figure \ref{fig:OP_Agar_results}) has quite different results from those seen in the task involving an experience buffer (Figure \ref{fig:ER_Agar_results}). All three algorithms start off climbing in performance as they do in the experience buffer task, but at around 100,000 training steps SPG's performance stagnates at a score of about 350 and CACLA's appears to plunge down to a performance resembling a random policy. DPG in the other had, seems to be unaffected by the lack of a buffer, as it manages to achieve the same performance as in the experience buffer task (Figure \ref{fig:ER_Agar_results}). 

\begin{figure}[!ht]
    \centering
    \includegraphics[width=.45\textwidth]{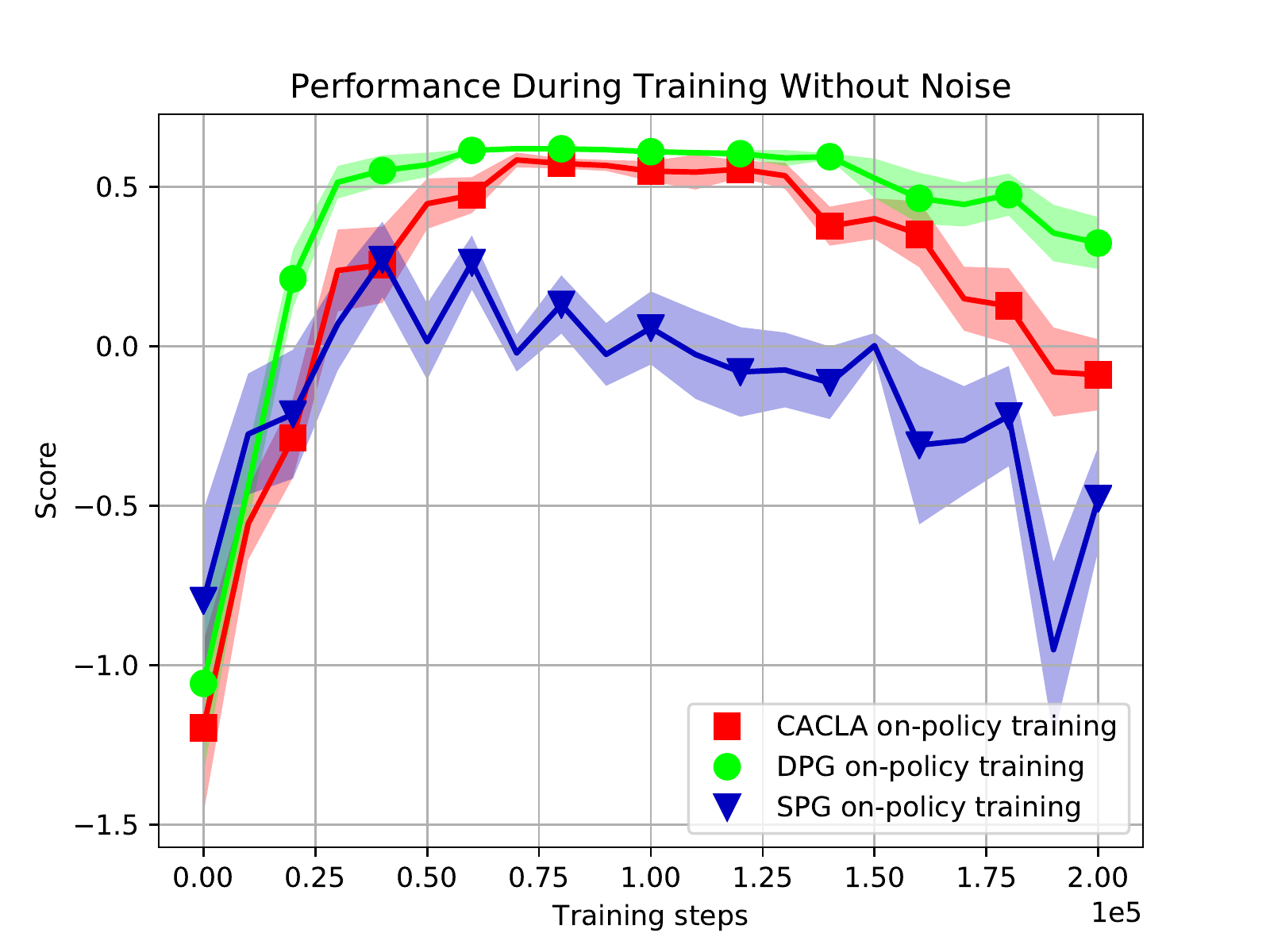}
    \caption{During-training performance in the task involving HalfCheetah-v2 without an experience buffer. Each point represents the average of the 10 testing rounds and the shaded area denotes its 1 S.D. range. Results are averaged over 10 simulations.}
    \label{fig:OP_HC_results}
\end{figure}

In the other had, in the on-policy training task in the HalfCheetah-v2 environment (Figure \ref{fig:OP_HC_results}), the worst performing algorithm appears to be SPG by a decent margin. The performance of all three algorithms appear to deteriorate over the training procedure, with the highest deterioration happening to SPG. CACLA also appears to deteriorate quite a lot, and while DPG also undergoes deterioration of performance, the amount is not as significant as in the other two algorithms. CACLA seems to, however, reach about the same max performance peak as DPG half way through the training procedure.

\subsection{Discussion}
After having given plenty of observation at how policies develop in the game of Agar.io, one hypothesis for the increased difficulty some algorithms had in some of the Agar.io tasks could be due to getting stuck in the corners of the map. One common behaviour for policies when the amount of noise begins to diminish is not being able to escape a corner. The map in Agar.io has no visible edge walls, as the area out of bounds appears the same to the agents as the regular playing field (except pellets do not spawn out of bounds). The agent thus has close to no visible feedback when a wall is keeping it from moving in a direction other than the pellets not moving relative to it. Given a small enough player size, the view range might even be too small for the agent to see pellets around him, thus moving causes no change in the state representation. Under a lack of noise in the agents actions, the player might spend long periods of time in a corner gathering data not useful towards learning (an episode is 2500 transitions in length). 

Given that for instance CACLA performs a policy update only when the observed return is higher than the predicted return, an agent using CACLA might spend lots of training steps not updating its policy in any meaningful way. This is specially the case when the agent is not using an experience buffer, which could explain the catastrophic forgetting observed in Figure \ref{fig:OP_Agar_results}. This phenomenon could also have caused DPG to have failed in two of training procedures in the Agar.io pixel input environment, leading to the lower average score with high SD (Figure \ref{fig:CNN_Agar_results}). Given that the experiment used much deeper actor networks, a policy that fails to start learning early would make its chances of eventually recovering be much lower due to the amount of action noise being less significant over the course of the training (as noise is the only reliable way to escape a corner for a poor policy). 

Lastly, the decay in performance that occurs on the on-policy training task of the HalfCheetah-v2 environment could be explained by a similar phenomenon. Possibly, as noise decreases over the training procedure, suboptimal transitions under the current policy that the agent might possibly not have yet learned to escape out of and were otherwise escaped through random noise, no longer are. In such a scenario, the policy progressively converges to this suboptimal local minima as there is no noise to allow it to escape it, which in turn, restricts the agent to only ever experiencing the same limited subset of the state-action space.

%% file: tex/conclusion.tex
\section{Conclusion}\label{sec:conclusions}
\subsection{Conclusion}
This study compared the new Sampled Policy Gradient (SPG) to two other well known actor-critic continuous action algorihtms, CACLA and DPG. The aim of this study was to extend the research performed in \cite{SPG} to other environments and task variations, as well as visit its performance with deep networks.

The results appear to be rather mixed. There was no task where SPG outperformed both of the algorithms, but SPG was also not consistently performaning worst. If anything, SPG held up rather well against the other algorithms in all the Agar.io tasks, although in the HalfCheetah tasks it underperformed. 

Surprisingly, despite CACLA being an on-policy algorithm, it was often the top performing of the three in the experience buffer tasks. In the other hand, in on-policy training tasks, it seemed to perform rather poorly, specially in the Agar.io environment. The reason behind this might have to do with its policy update rule only taking place in a subset of all transitions.

The hyperparameters used could have benefited from greater optimization, as they were only optimized for the Agar.io vision grid task. This could specially have affected the performance in the HalfCheetah-v2 tasks. The performance of DPG in the Agar.io pixel input task could be a manifestation of this, as DPG's hyperparameters are commonly claimed to be hard to optimize across different task \cite{DDPG}. As such, some of the results observed are to simply be viewed as a general idea of how the algorithms truly perform and compare.

\subsection{Future Work}

The approach taken by SPG of calculating the gradient from sampling the action space is somewhat reminiscent of the estimate of gradient performed by the REINFORCE algorithm \cite{REINFORCE}. In other techniques in machine learning, the numerical estimation of the gradient through sampling, although proven to be un-biased, has been shown to provide estimates of high variance \cite{RL:AI}. In Variational Autoencoders, a reparametrization trick is preferred over a sampled estimate of the gradient as it allows for a closed form solution to the gradient. Using the closed form gradient estimate in a high dimensional space is preferred over the sampled gradient due to needing exponentially larger samples to achieve the same sample density the higher the number of dimensions. The DPG algorithms is some sense makes use of this approach. It would thus be interesting to examine how the difference in performance between DPG and SPG varies as the dimensionality of the action space increases.

In this research, the amount of action samples per transition was set to 5. The manner in which SPG samples actions happens to not be scalable in an efficient manner for larger networks. Reason behind the lack of more deep network experiments in this investigation were due to the amount of training time for an individual run being in the order of days. SPG was the slowest of the algorithms, being around 10 times slower than CACLA and 2 times slower than DPG. SPG may thus be not well suited for tasks in deep reinforcement learning. In the other hand, given how quick modern GPUs can do forward propagation of large batches, experimenting with very large samples sizes in tasks where moderately sized networks are optimal, could shed light on the theoretical advantages in convergence that SPG might offer over DPG.

Furthermore, some of the runs in this research show SPG to have the greatest standard deviation in results out of the set of algorithms investigated. It would be interesting to determine to what extent the high standard deviation originates from the nature of the algorithm and how much arises from the fragility in its hyperparameter space.

%% file: tex/acknowledgements.tex
\section{Acknowledgements}
I would like to thank Herke van Hoof for supervising this project, as well as Marco Wiering and Anton Wiehe for their valuable feedback. 

Also, I would like to thank the Center for Information Technology of the University of Groningen for their support and for providing access to the Peregrine high performance computing cluster.

%% file: tex/appendix.tex
\section{Appendix}\label{ap:appendix}

\centering

\scalebox{1}{
    \begin{tabular}{l|l}
        \textbf{General parameters}                           & \textbf{Value}        \\ \hline
        Optimizer                                             & Adam                  \\
        Loss Function                                         & Mean-Squared Error    \\
        Weight Initializer                                    & Glorot Uniform        \\
        Hidden Layer Non-linearity                            & ReLU                 \\
        Prioritized Experience Replay Capacity                & 20,000                \\
        Training Batch Length                                 & 32                    \\
        Number Of Parallel Workers (On-policy task)           & 32                    \\
        Steps Between Target Network Updates                  & 1500                  \\
        Starting Standard Deviation                           & 1.0                   \\
        Standard Deviation At 50\% Training                   & 0.05                  \\
        CACLA Critic Learning Rate                            & 0.00075                \\
        CACLA Actor Learning Rate                             & 0.0005               \\
        DPG Critic Learning Rate                              & 0.0005                \\
        DPG Actor Learning Rate                               & 0.0001               \\
        SPG Critic Learning Rate                              & 0.0005               \\
        SPG Actor Learning Rate                               & 0.0001               \\
        Number Of Trainings Per Condition                     & 10               \\
        Testing Interval                                      & 5\%               \\
        Number Of Repetitions Per Test                        & 5               \\
    \end{tabular}
    }

\vspace{2.0cm}

\scalebox{1}{
    \begin{tabular}{l|l}
        \textbf{Agar.io vision grid parameters}               & \textbf{Value}        \\ \hline
        Episode length                                        & 2,500 training steps \\
        Frame Skip Rate                                       & 7                    \\
        Discount Factor                                       & 0.9                  \\
        Total Training Steps                                  & 500,000               \\
        Actor Hidden Layers                                   & 2                   \\
        Actor Units Per Hidden Layer                          & 100                   \\
        Actor Output Non-linearity                            & Sigmoid                \\
        Actor Output Units                                    & 2                \\
        Critic Hidden Layers                                  & 2                   \\
        Critic Units Per Hidden Layer                         & 100                   \\
        Critic Output Non-linearity                           & Linear                \\
        Experience Buffer Capacity                            & 40,000                \\
    \end{tabular}
    }
\clearpage
\scalebox{1}{
    \begin{tabular}{l|l}
        \textbf{Agar.io pixel input parameters}               & \textbf{Value}        \\ \hline
        Episode length                                        & 2,500 training steps \\
        Frame Skip Rate                                       & 7                    \\
        Discount Factor                                       & 0.9                  \\
        Total Training Steps                                  & 300,000               \\
        Convolutional Layers                                  & 2                   \\
        Input Image Size (Grayscale)                          & (42, 42, 1)           \\
        Convolutional Layer 1 Kernel Size                     & 8                   \\
        Convolutional Layer 1 Stride                          & 4                   \\
        Convolutional Layer 1 Filter Number                   & 32                  \\
        Convolutional Layer 2 Kernel Size                     & 4                 \\
        Convolutional Layer 2 Stride                          & 2                  \\
        Convolutional Layer 2 Filter Number                   & 64                  \\
        Actor Fully-connected Layers                          & 2                   \\
        Actor Fully-connected Layer Size                      & 100                  \\
        Actor Output Non-linearity                            & Sigmoid                \\
        Critic Fully-connected Layers                         & 2                   \\
        Critic Fully-connected Layer Size                     & 100                  \\
        Critic Output Non-linearity                           & Linear                \\
        Experience Buffer Capacity                            & 20,000                \\
    \end{tabular}
    }

\vspace{2.0cm}

\scalebox{1}{
    \begin{tabular}{l|l}
        \textbf{HalfCheetah-v2 parameters}                    & \textbf{Value}        \\ \hline
        Episode length                                        & 1,000 training steps \\
        Frame Skip Rate                                       & 0                   \\
        Discount Factor                                       & 0.99                 \\
        Total Training Steps                                  & 200,000               \\
        Actor Hidden Layers                                   & 2                   \\
        Actor Units Per Hidden Layer                          & 100                   \\
        Actor Output Units                                    & 6                \\
        Actor Output Non-linearity                            & Hyperbolic Tangent     \\
        Critic Hidden Layers                                  & 2                   \\
        Critic Units Per Hidden Layer                         & 100                   \\
        Critic Output Non-linearity                           & Linear                \\
        Experience Buffer Capacity                            & 15,000                \\
    \end{tabular}
    }